\documentclass{article}

\usepackage{PRIMEarxiv}

\usepackage[utf8]{inputenc} 
\usepackage[T1]{fontenc}    
\usepackage{hyperref}       
\usepackage{url}            
\usepackage{booktabs}       
\usepackage{amsfonts}       
\usepackage{amsmath}
\usepackage{amssymb}
\usepackage{nicefrac}       
\usepackage{microtype}      
\usepackage{lipsum}
\usepackage{fancyhdr}       
\usepackage{graphicx}       
\graphicspath{{media/}}     
\usepackage{authblk}
\usepackage[linesnumbered,ruled,vlined]{algorithm2e}
\usepackage[caption=false,font=footnotesize]{subfi
g}
\usepackage[backend=biber,style=ieee]{biblatex}
\SetKwInput{KwInput}{Input}                
\SetKwInput{KwOutput}{Output}              
\SetKwInput{KwInit}{Init} 
\SetKwInput{KwData}{Data}
\SetKwProg{Fn}{def}{\string:}{}

\fancypagestyle{firstpage}{%
    \lhead{\textit{ }}
  \lfoot{\scriptsize 978-1-6654-8045-1/22/\$31.00 \copyright2022 IEEE}
}

\title{Autonomous Drug Design with Multi-Armed Bandits
}

\author[1,2]{Hampus Gummesson Svensson\thanks{Corresponding author: \href{mailto:hamsven@chalmers.se}{hamsven@chalmers.se}}\hspace{1.5mm}\textsuperscript{,}}
\author[2]{Esben Jannik Bjerrum\thanks{This author is currently at Odyssey Therapeutics, Cambridge, MA, USA.}\hspace{1.5mm}\textsuperscript{,}}
\author[3]{Christian Tyrchan}
\author[1,2]{Ola Engkvist}
\author[1]{Morteza Haghir Chehreghani}

\affil[1]{Chalmers University of Technology, Department of Computer Science and
Engineering, Gothenburg, Sweden}
\affil[2]{AstraZeneca, Molecular AI, Discovery Sciences, R\&D, Gothenburg, Sweden}
\affil[3]{AstraZeneca, Medicinal Chemistry, Research and
Early Development, Respiratory and
Immunology (R\&I),
BioPharmaceuticals R\&D,
Gothenburg, Sweden}

\begin{document}
\maketitle
\thispagestyle{firstpage}
\begin{abstract}
  Recent developments in artificial intelligence and automation support a new drug design paradigm: autonomous drug design. Under this paradigm, generative models can provide suggestions on thousands of molecules with specific properties, and automated laboratories can potentially make, test and analyze molecules with minimal human supervision.  However, since still only a limited number of molecules can be synthesized and tested, an obvious challenge is how to efficiently select among provided suggestions in a closed-loop system. We formulate this task as a stochastic multi-armed bandit problem with multiple plays, volatile arms and similarity information. To solve this task, we adapt previous work on multi-armed bandits to this setting, and compare our solution with random sampling, greedy selection and decaying-epsilon-greedy selection strategies. According to our simulation results, our approach has the potential to perform better exploration and exploitation of the chemical space for autonomous drug design.
\end{abstract}


\section{Introduction}
\begin{figure}[h!]
    \centering
    \includegraphics[width=0.9\textwidth]{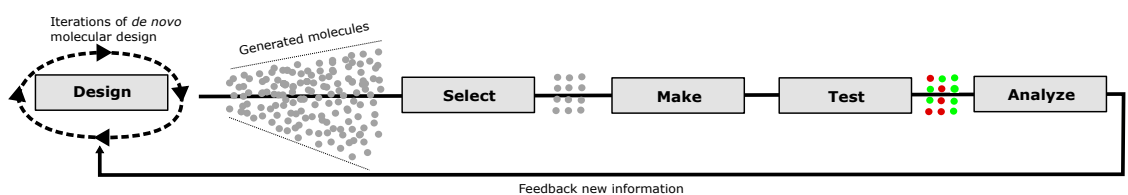}
    \caption{A schematic illustration of the (autonomous) drug design process.}
    \label{fig:dsmt}
\end{figure}
Developing a new drug is a complex process that can take up to a decade and cost more than US $\$1$ billion \cite{wouters2020estimated}. A crucial part of this process is to design novel clinical drug candidates with desired molecular properties \cite{hughes2011principles}. Drug candidates are usually identified in an iterative optimization process consisting of four steps, the so-called Design-Make-Test-Analyze (DMTA) cycle. In the design step, molecules are designed to have some specific properties, including a high binding affinity to a specific protein. In the make step, it is decided how to synthesize them. After being synthesized, the properties of the molecules are experimentally measured in the test step. Subsequently, in the analyze step the newly gathered experimental data is used to improve the design choices in the next round. 
The time it takes to complete a cycle is a major factor in the overall productivity and, therefore, one wants to both reduce cycle times and minimize the number of cycles needed to find and optimize the properties of drug candidates. 

Two new paradigms have emerged to increase the productivity in drug design: \emph{de novo} molecular design and accelerating the DMTA cycle through automation. Recent advances in \emph{de novo} molecular design utilize generative models for sampling the chemical space and in this way generate molecules with specific desirable properties  \cite{meyers2021novo, gao2022sample}.
The automation of the DMTA cycle takes advantage of automated laboratories and machine learning to make, test and analyze molecules with minimal human intervention \cite{coley2020autonomous1, coley2020autonomous2, hase2019next, stein2019progress, shen2021automation}.
Used together, these technologies can enable \emph{autonomous drug design}. Under such paradigm, \emph{de novo} molecular design is used to generate an extensive list of molecules using limited prior information. However, it is only possible to synthesize a fraction of the proposed molecules, as illustrated in \figurename~\ref{fig:dsmt}, since each experiment is both costly and time-consuming. Even if it would be possible to synthesize all generated molecules, it can be more efficient to obtain new information sequentially in an adaptive manner. Hence, the autonomous drug design system needs to decide, with minimal human supervision, on which of the designed molecules to make. Subsequently, the automated laboratory tries to synthesize the selected molecules and, if successful, measure their properties. Using the newly acquired information, the system updates its knowledge to better steer the molecular \emph{de novo} design towards desired areas of the chemical space.  

This work is focused on how to select  molecules to make in order to explore and exploit the chemical space efficiently. In each iteration, we obtain a list of molecules that can be selected. The drug-like chemical space, and thus the number of selectable molecules, has been estimated to be up to $10^{60}$ molecules \cite{reymond2015chemical}. Each molecule has a feature vector in terms of descriptor(s) that encodes its chemical and/or structural characteristics, e.g., the Morgan fingerprint \cite{morgan1965generation} which is a popular way to encode structural features of molecules in a bit or count vector. Moreover, when utilizing \emph{de novo} molecular design to suggest promising molecules, the set of suggested molecules can be different in each iteration, which makes it even more challenging. 
Compared to other works that have focused on \emph{de novo} molecular design \cite{blaschke2020reinvent, mercado2020graph, atance2022graphinvent} or automating the synthesis of molecules \cite{christensen2021data}, we simulate all steps of the DMTA cycle by creating a digital twin. Previous work by \textcite{matveieva2021benchmarks} has shown that predictive models trained on structural features of molecules are able to learn a ground truth determined by pre-defined patterns, e.g., number of nitrogen.
Given a ground truth that simulates the test scores of a desired target, this enables us to investigate how the choices in one cycle affect the succeeding cycles. 

We study this problem in the context of the multi-armed bandit (MAB) problem \cite{lattimore2020bandit,slivkins2019introduction}, where the goal is to adaptively compute the most informative decisions. In the original problem introduced by \cite{lai1985asymptotically}, a decision-maker must choose from $M$ possible actions, so-called \emph{arms}, for $T$ rounds. In each round, the decision-maker chooses a base arm and observes a reward for this arm, which is drawn independently from a fixed distribution that is not known to the decision-maker. The objective is to maximize the expected cumulative reward, by identifying the arm with the highest expected reward in each round, while choosing as few suboptimal arms as possible. Many extensions of this problem have been developed to fit different problem settings. One such extension is MAB with similarity information, where each arm corresponds to objects with feature vectors. This makes it possible to measure the similarity between the arms by computing the distance between the feature vectors. 

Considering the problem at hand, we regard the following characteristics of our stochastic MAB problem: (1) multiple base arms are played in each round, corresponding to choosing a super arm of several distinct base arms in each round; (2) base arms are volatile, meaning that in each round, the set of available base arms to choose from may change; (3) the expected outcome of a base arm depends on its feature vector; (4) the set of possible feature-arm pairs is in practice infinite, since there are up to $10^{60}$ possible molecules and feature vectors of 2048 bits gives $2^{2048}$ possible feature vectors.
To solve this problem adequately, we propose to extend the contextual Zooming algorithm \cite{slivkins2011contextual} to our problem. We use their proposed techniques to allow for volatile base arms and extend it further to enable multiple plays of arms in each round. Also, instead of using contextual information, we define the base arms by their structural feature vectors, which together with Jaccard distance provides a natural metric space, to adaptively create a partition of the arm space.
To the best of our knowledge, this is the first study that simulates an automated DMTA cycle and, in this setting, investigates a multi-armed bandits approach to determine which molecules to make next. 

The rest of the paper is organized as follows.
Firstly, in Section \ref{sec:related_work} we discuss the related work to our stochastic MAB problem and the existing methods. In Section \ref{sec:problem} we formulate our problem as a stochastic multi-armed bandits problem. Then, in Section \ref{sec:select_strategies} we present our extension to the contextual Zooming algorithm. Next, in Section \ref{sec:experiment}, we describe the experimental results, and finally in Section \ref{sec:conclusion}, we conclude the paper.

\section{Related Work}
\label{sec:related_work}
The multi-armed bandit (MAB) framework provides a principled way to model the exploration/exploitation trade-off for sequential decision-making under certainty.
It has been widely used in different applications such as medical trials \cite{Villar2015,Press09}, news recommendation \cite{LiCLS10}, finance \cite{ShenWJZ15}, navigation  \cite{Akerblom0C20} and bottleneck identification \cite{OnlineMinimax}.

In this paper, we consider the stochastic MAB problem, where the reward for each arm is drawn independently from a fixed but a priori unknown distribution.  Hereafter, stochastic MABs are referred to as simply MABs. Here we present related work for our proposed approach, in particular MAB with similarity information. We direct the attention to \textcite{slivkins2019introduction} and \textcite{lattimore2020bandit} for a comprehensive overview of different extensions, and corresponding algorithms, of the original MAB problem.

\subsection{Contextual MAB}
The contextual MAB problem has been broadly studied under the linear realizability assumption, introduced by \textcite{abe2003reinforcement}, where the expected reward is assumed to be linear with respect to a feature vector of each arm \cite{chu2011contextual,agrawal2013thompson,li2010contextual,auer2002using}. There has been a great success in using the contextual MAB problem to model real-life applications, such as recommender systems, health applications and information retrieval \cite{bouneffouf2020survey}. 

\subsection{Multiple-Play MAB}
The original MAB formulation introduced by \textcite{lai1985asymptotically} considers single plays, where $K=1$ arm is chosen in each round \cite{agrawal1990multi, komiyama2015optimal}. The extension of choosing $K > 1$ arms in each round was introduced by \textcite{anantharam1987asymptotically}. This is a special case of the combinatorial MAB problem \cite{chen2013combinatorial,gai2012combinatorial}, where an allowed combination of several arms is played at each round. Whereas, in the multiple-play MAB setting, all combinations of $K$ arms are allowed and a reward is observed for each individual base arm.

\subsection{MAB with Volatile Base Arms}
A usual assumption is that all arms are available at each round. However, in many applications, including ours,  this is  not the case. For instance, a molecule suggested by the designer in the current round may not be available in the next round, due to its properties not being of interest anymore. Also, even though testing molecules yields an inherent uncertainty, it is a waste of resources to test molecules multiple times, meaning that tested molecules should not be available in the coming rounds. \textcite{kleinberg2010regret} study \emph{sleeping bandits}, where the set of available actions is allowed to vary adversarially from one round to the next. They assume a fixed finite number of arms and a stochastic adversary. They propose an algorithm that prioritizes playing an arm that has become available for the first time. Otherwise, it plays the arm with the largest upper confidence bound, inspired by the algorithm UCB1 \cite{auer2002finite}.

\subsection{MAB with Similarity Information}
Although a large set of MAB algorithms have been proposed in the literature with a fixed small number of arms, MAB problems with infinite or exponentially large arm sets are still actively studied. For such setting, one common approach is to use similarity information between contexts and/or arms, by assuming that similar actions yield similar qualities. 

\textcite{kleinberg2008multi} introduce the \emph{Zooming algorithm}, where the similarity information is given as a metric space of arms \cite{kleinberg2019bandits}. Their algorithm tries to approximately learn the expected rewards over the metric space by probing different ``regions'' of the space, which leads to an adaptive partitioning of the metric space \cite{slivkins2019introduction}. At each round $t$, there is a set of active arms, determined by an activation rule. Each active arm $x$ covers a region of the metric space. This region is given by the confidence ball of the arm $B(x,r_t(x))$, which is a ball with the arm at its center. The radius of the ball is the confidence radius $r_t(x)$ of the empirical average reward (of the active arm) at round $t$. The confidence radius is related to the size of the one-sided confidence interval of the empirical average reward and guarantees, with high probability, that the difference between the true expected reward and empirical average reward is not more than the confidence radius. To determine what active arm to play, it chooses an arm with the largest upper confidence bound, similar to arm selection of algorithm UCB1 \cite{auer2002finite}. 

\textcite{slivkins2011contextual} extends the Zooming algorithm to the contextual setting, where the similarity information is given as a metric space of context-arm pairs. Our works extend the techniques developed in this work to allow volatile arms and multiple plays. We relax the contexts and define the arm-space by the corresponding feature vectors.

\textcite{bubeck2010x} consider stochastic MAB where the arm set can be a generic measurable space, allowing infinite arm sets. They assume the existence of a dissimilarity function that constraints the expected reward function. However, their formulation does not allow multiple plays and volatile base arms. They propose an algorithm that adaptively discretizes the arm set by maintaining a binary tree whose leafs are associated with measurable regions of the arm-space. They assume  \emph{a priori} choice of a covering tree, which may be difficult to construct for non-standard metric spaces.

\textcite{chen2018contextual} study contextual combinatorial MAB with volatile arms and submodular rewards. This allows selection of several arms in each round, and the set of available arms to vary in each round. They use a greedy oracle to, in each round, return an approximation of the best super arm of available base arms. Moreover, their proposed algorithm utilizes similarity information given as a fixed discretization of the context space, by partitioning the context space into hypercubes of identical size. A fixed discretization is limited in how much it can learn an arbitrary structure of the expected reward function in the similarity space, and it may be preferable to adaptively learning the structure.

\textcite{nika2020contextual} also consider contextual combinatorial MAB with volatile base arms, but without the assumption of submodular rewards. They introduce an algorithm that adaptively discretizes the context space using similarity information, given as a well-behaved Euclidean metric space. The adaptive discretization utilizes a tree of partitions, where the algorithm maintains an active set of leaf nodes whose regions cover the context space. Moreover, their algorithm uses an approximation oracle to select a super arm of base arms in each round.

\section{Problem Formulation}\label{sec:problem}
\begin{algorithm}[h!]
\DontPrintSemicolon
\KwInput{Dissimilarity space $(\mathcal{X}, \mathcal{D}_{\mathcal{X}}$)}
\For{each round $t=1,\dots,T$}{
    $M^t > K$ base arms, indexed by the set $\mathcal{M}^t$, arrive with corresponding feature vectors $\mathcal{X}^t \subset \mathcal{X}$ and scores $\mathcal{F}^t \in \left[0,1\right]^{M^t}$\;
    Choose super arm $\mathcal{S}^t \subset \mathcal{M}^t$ of $K$ distinct base arms\;
    Observe reward $r(x^t_m) \in \left\{0,1 \right\}$, $\forall m \in \mathcal{S}^t$\;
    }
\caption{MAB formulation of the autonomous drug design task} \label{alg:problem_formulation}
\end{algorithm}

We formulate the autonomous drug design task as a stochastic multi-armed bandits problem proceeding over $T$ rounds, summarized in Algorithm \ref{alg:problem_formulation}, by extending the problem formulations of \textcite{nika2020contextual} and \textcite{slivkins2011contextual} to adapt to the specifications of our task. Each base arm is defined by a $D$-dimensional feature vector $x$, belonging to the arm space $\mathcal{X}\subseteq \mathbb{R}^D$. Moreover, for each pair of base arms $x,x' \in \mathcal{X}$, we have a dissimilarity function
$0 \leq \mathcal{D}_{\mathcal{X}}(x,x') \leq 1$ that measures the dissimilarity between them. The arm space and dissimilarity function create a dissimilarity space $(\mathcal{X}, \mathcal{D}_{\mathcal{X}}$). In our problem, arms are molecules, where the feature vectors are molecular fingerprints and the dissimilarity function measures the chemical dissimilarity. The stochastic outcome of a base arm with feature $x$ is denoted by $R(x)$.  In this work, we restrict to the setting with Bernoulli rewards, where rewards for each feature are given by a Bernoulli random variable that takes values in $\{0,1\}$. We assume that there exists an unknown function $\mu(x): \mathcal{X} \rightarrow \left[0,1\right]$ such that $\mu(x) = \mathbb{E}[R(x)],$ $\forall x \in \mathcal{X}$. In this work, we let this unknown function be determined by a ground truth that simulates our desired target.

At round $t$, a set of $M^t > K$ base arms, indexed by the set $\mathcal{M}^t$, arrive with corresponding feature vectors $\mathcal{X}^t$ and scores $\mathcal{F}^t$ such that $x^t_m$ is the feature vector of base arm $m \in \mathcal{M}^t$ in round $t$. In this work, when only considering Bernoulli rewards, the score $f^t_m$ of base arm $m$ at round $t$ is the estimated probability of $\mu(x_m^t) > 0.5$, i.e., the probability that the molecule with feature vector $x_m^t$ binds to the target protein. This score is provided by a machine learning model, called hereafter the \emph{scoring function}, and is trained before arms arrive. After the base arms, features and scores have arrived, a selection strategy is used to select a super arm $\mathcal{S}^t \subset \mathcal{M}^t$ of $K$ base arms. Next, $\mathcal{S}^t$ is performed and for each base arm $m\in \mathcal{S}^t$ a reward $r(x^t_m)$ is observed, i.e., a realization of the random variable $R(x^t_m)$. The objectives are to maximize the cumulative reward by round $T$ and find a diverse set of base arms with high rewards. This corresponds to finding a set of \emph{novel} drug candidates by round $T$.

\section{Zooming with Multiple Plays and Volatile Arms}
\label{sec:zooming}

\begin{algorithm}[H]
\DontPrintSemicolon
  \KwInput{Dissimilarity space $(\mathcal{X}, \mathcal{D_{\mathcal{X}}})$ of diameter $\leq 1$}
  \KwData{Collection of initial history $\mathcal{H}_0$ consisting of plays of initial base arms and corresponding rewards.}
  \KwInit{$B \gets Ball(p,1)$; Add initial history $\mathcal{H}_0$ to $B$; $\mathcal{A} \gets B$ }
  \For{each round $t = 1,\dots, T$}{
    Observe base arms in $\mathcal{M}^t$ and feature vectors $\mathcal{X}^t$\;
    $\mathcal{R}^t \gets \{B \in \mathcal{A}: \exists m\in \mathcal{M}^t, m \in \texttt{dom}(B, \mathcal{A})\}$\;
    Compute indices $g^t(B)$, $\forall B \in \mathcal{R}^t$\;
    Compute indices $g^t(x_m^t)$, $\forall m\in \mathcal{M}^t$\;
    $\mathcal{S}^t \gets \texttt{Oracle}(g^t(x_1^t), \dots,g^t(x_{M^t}^t))$\;
    Observe reward $r(x^t_m)$ for each base arm $m\in\mathcal{S}^t$\;
    \tcp{Update counters for each ball}
    \For{$m \in \mathcal{S}^t$}{
        $n(B^t_{m}) \gets n(B^t_{m}) + 1$\;
        $\texttt{rew}(B^t_{m}) \gets \texttt{rew}(B^t_{m}) + r(x^t_{m})$\;
    }
    \For{all distinct $B \in \{B_{m}^t: m\in  \mathcal{S}^t\}$}{
        \If{$\texttt{conf}(B) \leq \texttt{radius}(B)$}{
            \tcp{Refine partition}
            $m' \gets \text{argmax}\{r(x_m^t): m \in \mathcal{S}^t, B^t_{m} = B\}$\;
            $B' \gets Ball(x_{m'}^t, \frac{1}{2}\texttt{radius}(B))$\;
            Update $n(B')$ and $\texttt{rew}(B')$ using relevant history\;
            $\mathcal{A} \gets \mathcal{A} \cup \{B'\}$\;
        }
    }
  }
\caption{Zooming with multiple plays and volatile arms.} \label{alg:zooming}
\end{algorithm}

In this section, we introduce our method \emph{Zooming with multiple plays and volatile arms}, summarized in Algorithm \ref{alg:zooming}. It is an extension of the contextual Zooming algorithm \cite{slivkins2011contextual} to our problem, as formulated in Section \ref{sec:problem}. Compared to the contextual Zooming algorithm, we relax the contexts and instead let each arm be fully defined by its feature vector. We employ their  methods for dealing with volatile base arms. Also, we enable multiple plays by using a greedy oracle that selects a super arm based on the index of each base arm.

In each round $t$, there is a set of activated balls $\mathcal{A}$, where each ball $B \in \mathcal{A}$ constitutes a ball with radius $\texttt{radius(B)}$ in the dissimilarity space.
After base arms $\mathcal{M}^t$ and feature vectors $\mathcal{X}^t$ have been observed, the set of relevant balls $\mathcal{R}^t$ is determined. A ball $B \in \mathcal{A}_t$ is relevant if its \emph{domain} covers at least one of the base arms in $\mathcal{M}^t$.
The \emph{domain} of ball $B$ is a subset of $B$ that excludes all balls $B' \in \mathcal{A}_t$ with smaller radius $\texttt{radius} (B')$
\begin{equation}
    \texttt{dom}_t (B) \triangleq B \setminus \left(\cup_{B'\in \mathcal{A}_t: \texttt{radius}(B')< \texttt{radius}(B)} B' \right).
\end{equation}

For each relevant ball $B \in \mathcal{R}^t$, we calculate its index as defined by \textcite{slivkins2011contextual}
\begin{equation}
    g^t(B) \triangleq \texttt{radius}(B) + \text{min}_{B' \in \mathcal{A}_t} \left(I_t^{\texttt{pre}} (B') + \mathcal{D} (B,B') \right), 
\end{equation}
where $\mathcal{D} (B,B')$ is the dissimilarity between the centers of the two balls. The \emph{pre-index} $I_t^{\texttt{pre}}$ of ball $B$ is defined by
\begin{equation}
     I_t^{\texttt{pre}}(B) \triangleq \nu_t (B) + \texttt{radius}(B) + \texttt{conf}_t(B),
\end{equation}
where $\nu_t (B) \triangleq \frac{\texttt{rew}_t(B)}{\text{max}(1,n_t (B))}$ is the average reward of ball $B$, given by the total reward $\texttt{rew}_t(B)$ and total number of plays $n_t (B)$ of ball $B$. Moreover, $\texttt{conf}_t(B)$ is the confidence radius of ball $B$ at time $t$, given by 
\begin{equation}
    \texttt{conf}_t(B) \triangleq 4 \sqrt{\frac{\log T}{1 + n_t(B)}}.
\end{equation}
Given the indices of all relevant balls, we want to compute indices of all base arms in $\mathcal{M}^t$. We investigate two methods to determine the indices of the base arms.  The first method computes the indices by
\begin{equation}
    g^t(x_m^t) = f^t_m \times g^t(B_m^t), \forall m\in \mathcal{M}^t,
\end{equation}
where $B_m^t\in \mathcal{R}^t$ is the relevant ball covering base arm $m$ at round $t$, i.e., the domain of ball $B_m^t$ covers base arm $m$ at round $t$. We call this method weighted Zooming with multiple plays and volatile arms (weighted Zooming in short). The second method computes the indices by
\begin{equation}
    g^t(x_m^t) = g^t(B_m^t), \forall m\in \mathcal{M}^t.
\end{equation}
We call this variant unweighted Zooming with multiple plays and volatile arms (unweighted Zooming in short).
Given the indices of each base arm, we want to select a super arm $\mathcal{S}^t$ consisting of $K$ base arms. This is done using an oracle that selects the set of $K$ base arms with the largest cumulative index and breaks ties arbitrarily, as illustrated in Algorithm \ref{alg:oracle}. 
A reward is observed for each base arms in the super arm $\mathcal{S}^t$, and subsequently the total number of observations $n$ and total reward $\texttt{rew}$ are update for the balls covering these base arms. If the confidence radius of a ball $B$ is less than or equal to its radius, the partition is refined by creating a new ball $B'$ with half the radius of $B$. The center of the new ball $B'$ is the feature vector of the base arm with the largest reward that was selected in the current round and is covered by ball $B$. For the new ball $B'$, we add all relevant history from parent $B$, meaning that we update the counters using the previously observed rewards of base arms in the \emph{domain} of $B'$. Lastly, we add the new ball $B'$ to the set of activated balls $\mathcal{A}$.

If the time horizon $T$ is not known in advance, a common technique is to divide the rounds into phases, and at the beginning of each phase incrementally increase the time horizon, e.g., by geometric doubling where the time horizon is given by $T_i = ar^i$ for phases $i=0,1,2,\dots$, $a\in \mathbb{R}_{>0}$ and $r\in \mathbb{R}_{>1}$. This is called the \emph{doubling trick}. When a multi-armed bandit algorithm is also restarted at the beginning of each phase, it has been shown that some doubling tricks can enjoy certain performance guarantees established for a fixed time horizon \cite{besson2018doubling}.

\begin{algorithm}[H]
\DontPrintSemicolon
\Fn{$\texttt{Oracle}(g_1, \dots, g_{M})$}{
    $\mathcal{S} \gets \text{max}_{\mathcal{C} \subseteq \{1, \dots, M\}: |\mathcal{C}| = K} \sum_{i\in C} g_i $\;
    \textbf{return} $\mathcal{S} $\;
}
\caption{Oracle} \label{alg:oracle}
\end{algorithm}

\begin{figure}[h]
     \centering
     \subfloat[Normalized cumulative reward]{
    \includegraphics[width=0.49\textwidth]{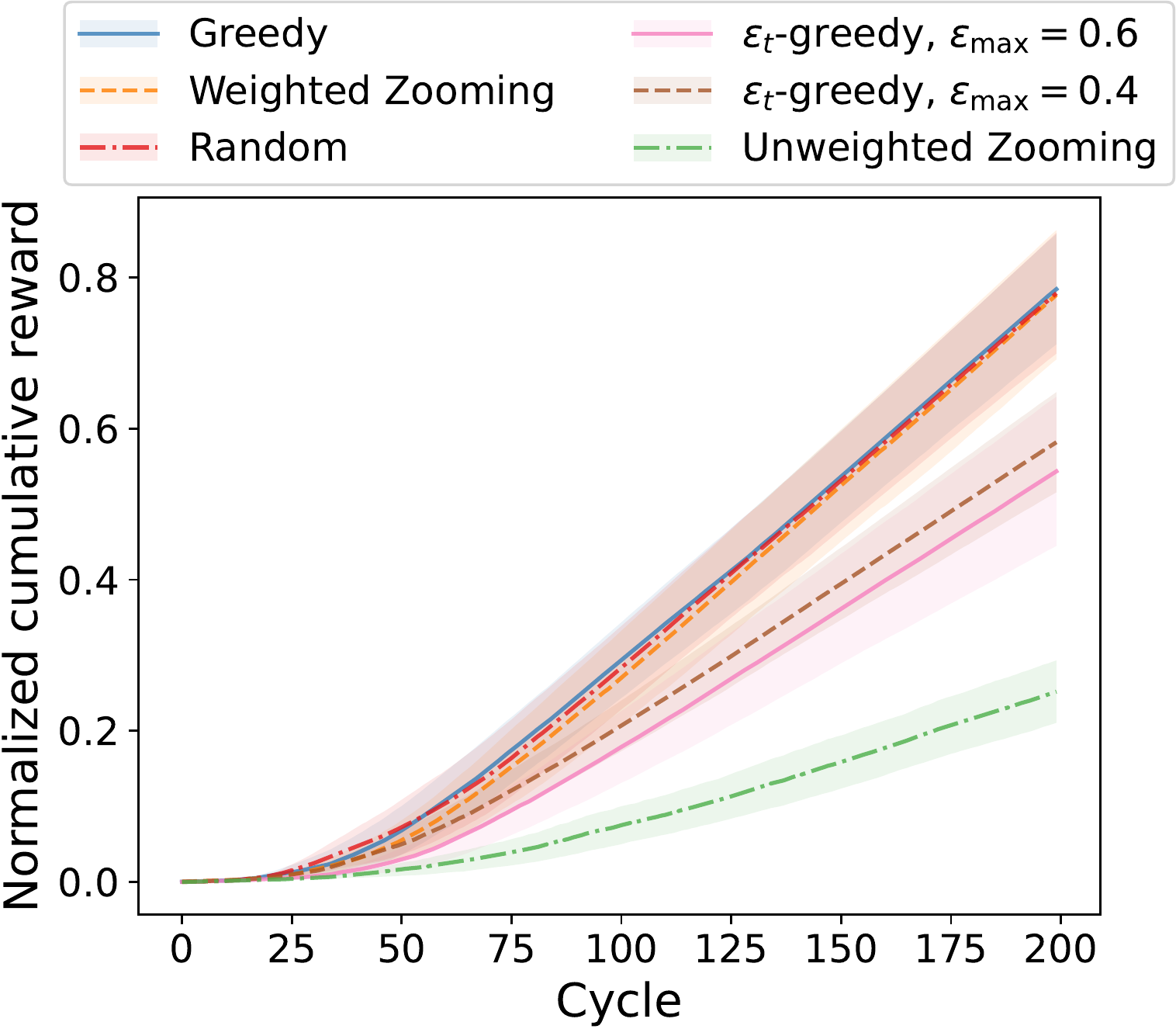}\label{fig:cum_rew}}
     \hfill
    \subfloat[Novelty of selected actives]{\includegraphics[width=0.49\textwidth]{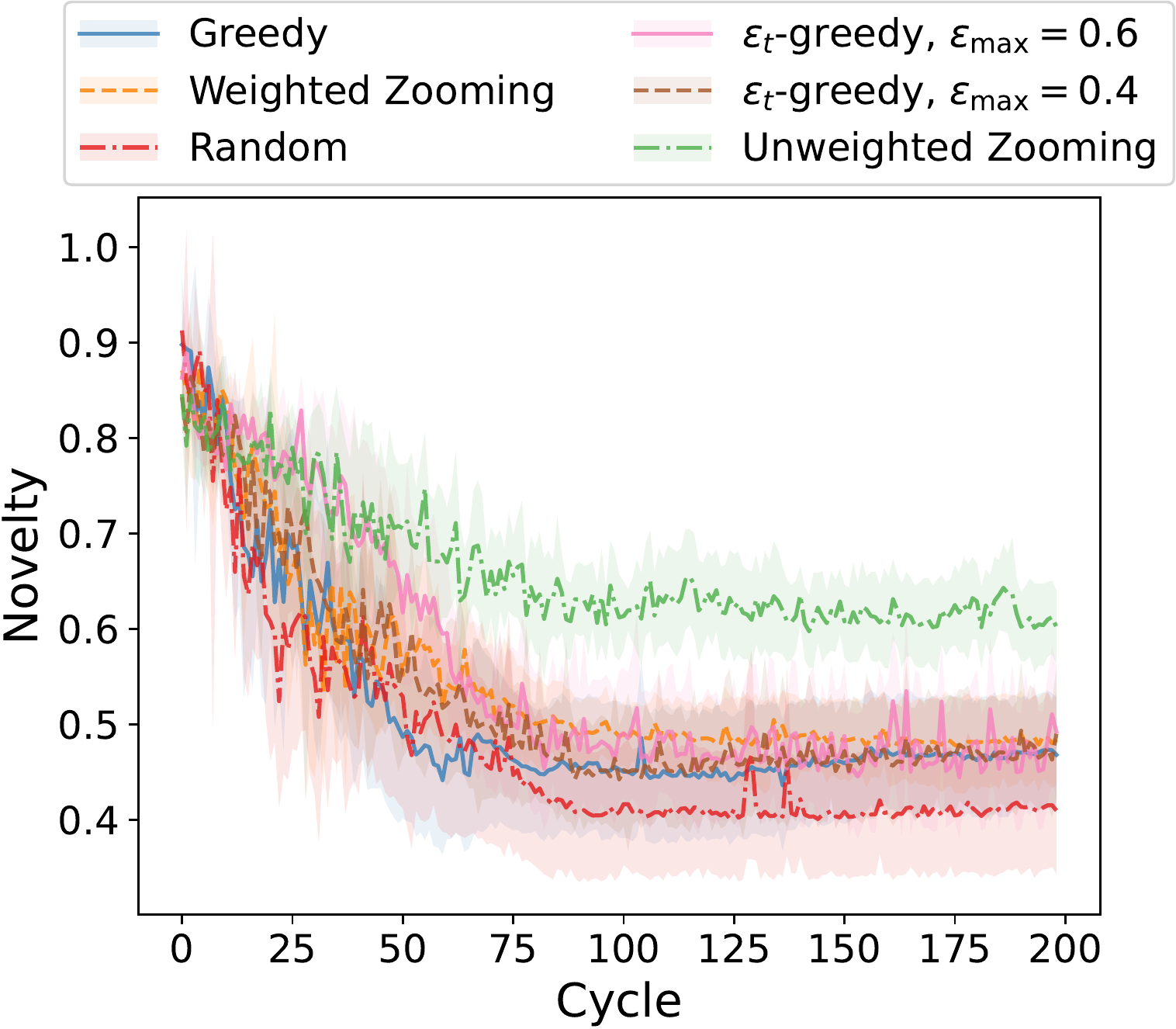}\label{fig:novelty}}
     \hfill
     \subfloat[Mean of cum. reward and novelty]{\includegraphics[width=0.49\textwidth]{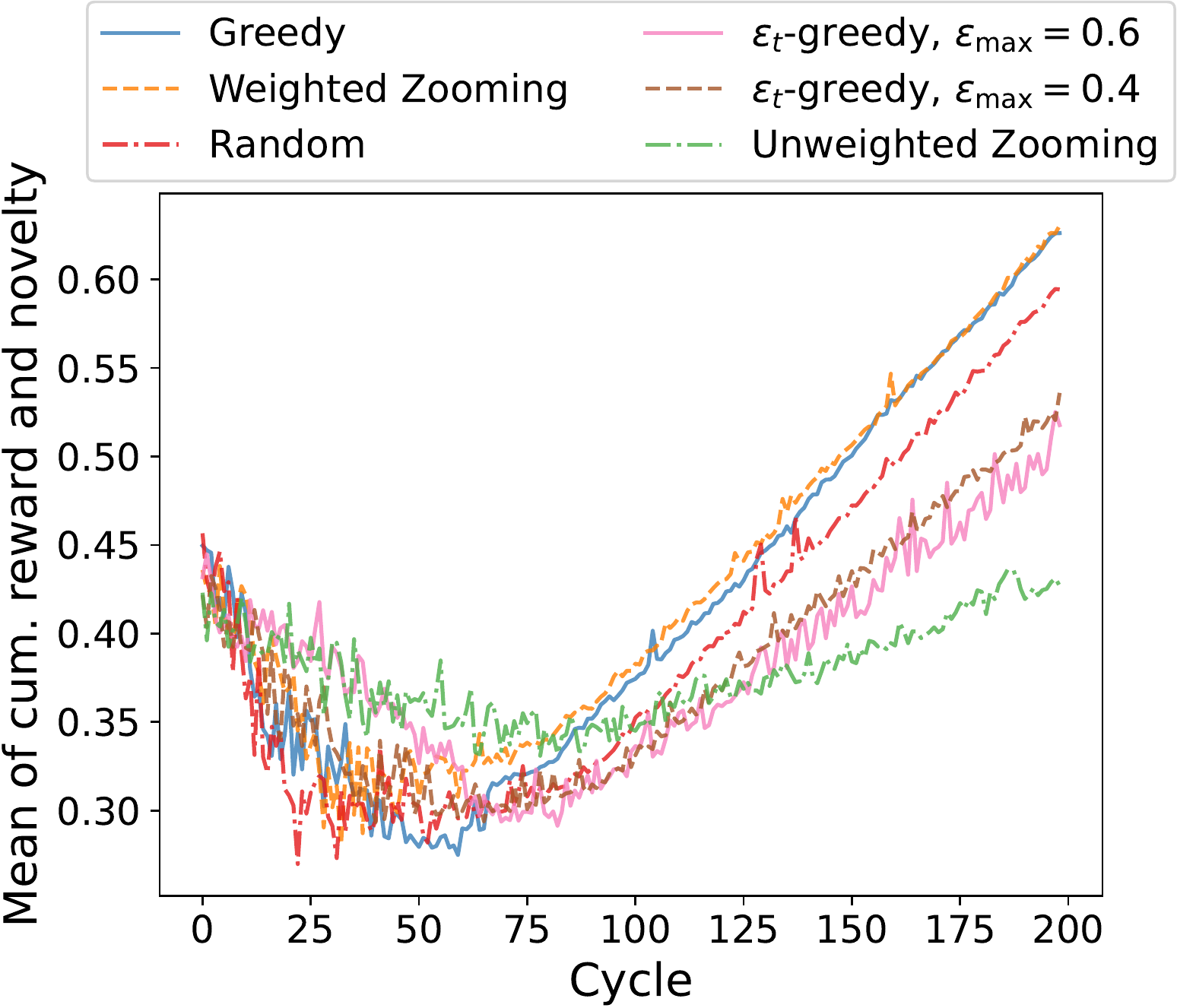}\label{fig:mean_rew_nov} }
    \caption{Normalized cumulative reward, novelty of selected actives and the mean of the former two averaged over 10 runs for each selection strategy. For the former two, the 95\% approximate confidence intervals of the averages over 10 runs is shown. A novelty of 1 corresponds to selecting actives that are entirely dissimilar to previously selected actives, while a novelty of 0 corresponds to a selection that is equal in similarity to the previously selected actives. $\epsilon_t$-greedy with $\epsilon_{\text{max}} = 0.6$ and unweighted Zooming show good performance with regard to both cumulative reward and novelty in the first 50 cycles, while weighted Zooming and greedy both performs well for at least the last 100 cycles.}
    \label{fig:results}
\end{figure}

\section{Experiments}
\label{sec:experiment}

\begin{algorithm}[H]
\DontPrintSemicolon
\For{each cycle (round) $t=1,\dots,200$}{
    $M^t$ molecules (base arms) are generated as described in Section \ref{sec:simulation_DMTA}\;
    Compute Morgan fingerprints (feature vectors $\mathcal{X}^t$)\;
    Select a set $\mathcal{S}^t$ of $K=100$ molecules (base arms) using selection strategy\;
    Simulate make, test and analyze as described in Sections \ref{sec:simulation_DMTA}\;
    Observe activity (reward) of each selected molecule (base arm)\;
    If applicable, update and improve selection given observations\;
    Improve scoring function given observations\;
    }
\caption{Experimental procedure of DMTA cycle} \label{alg:experimental_procedure}
\end{algorithm}

Algorithm \ref{alg:experimental_procedure} describes the different steps of our experimental procedure.
We investigate various strategies that in each round $t$ select a set of $\mathcal{S}^t$ of molecules, i.e., a super arm of base arms. Lines 2-4 and 6-7 of Algorithm \ref{alg:experimental_procedure} corresponds to lines 2-6 and 7-16 of Algorithm \ref{alg:zooming}, respectively.
For each selection strategy, we perform 10 runs, where each run corresponds to 200 sequential cycles of a simulated DMTA cycle or in other words $T=200$ rounds of the stochastic multi-armed bandit problem formulated in Section \ref{sec:problem}. As feature vectors for each base arm, we use the 2048-bit Morgan fingerprints \cite{morgan1965generation} with radius 2. They are generated with RDKit \cite{landrum2006rdkit} using feature-based invariants and counts of structural features. To measure the dissimilarity between two features, we use the Jaccard distance, which fulfills all properties of a metric \cite{kosub2019note}. This means that the similarity information is given as a metric space.
Below, we describe the simulated DMTA cycle, investigate selection strategies and demonstrate the corresponding results.

\subsection{Simulation of DMTA Cycle}
\label{sec:simulation_DMTA}
\paragraph{Design}
    To generate chemical molecules in each cycle, we use the \emph{de novo} molecular design tool REINVENT \cite{blaschke2020reinvent}, which is an approach based on SMILES \cite{weininger1988smiles} and uses reinforcement learning for learning. Here, we provide details on the hyperparameters that we use, but we refer to the work of \textcite{blaschke2020reinvent} for more details on the generation and hyperparameters. In each iteration of REINVENT, a batch of 128 molecules are generated and, subsequently, scored by a scoring function. The score of each molecule is the predicted probability of it being active, according to the ground truth. To improve the generation, each score is fed into the learning loop without any transformation. REINVENT performs at least 500 iterations of generation, and then we use the following stopping criteria.  The mean score of each batch is calculated and if the highest mean score, over all previous iterations, is not improved over 50 iterations in a row, REINVENT stops. Moreover, REINVENT is used with the identical Murcko scaffold diversity filter with a bucket size of 100, minimum score of 0.2 and minimum similarity of 0.6. 
    
    As scoring function, which gives probability that the molecule with feature vector $x_m^t$ binds to the target, we use a quantitative structure-activity relationship (QSAR) model based on a random forest model. 
    The initial QSAR model is trained on 20 active and 100 inactive molecules, with respect to the ground truth, that were randomly sampled from ChEMBL \cite{gaulton2012chembl} using REINVENT.
\paragraph{Make}
    In this work, we assume that all molecules are possible to make, and all molecules are equally time-consuming and yield the same cost to make.
\paragraph{Test}
    The test step provides noisy test scores of every molecule. It consists of a ground truth providing true test scores and a noise model that adds noise to these scores. We use binary scores where a molecule is either inactive (0) or active (1) with respect to a target modelled by the ground truth. Previous work by \textcite{matveieva2021benchmarks} has shown that QSAR models are able to learn scores determined by pre-defined patterns. Following this work, the score of each molecule is determined by the following ratios between counts of carbon $n_{\text{c}}$, nitrogen $n_{\text{n}}$ and oxygen $n_{\text{o}}$ atoms
    \begin{align}
        5.5& \leq  \frac{n_{\text{c}}}{n_{\text{o}}} \leq 5.67,\\
        7& \leq \frac{n_{\text{c}}}{n_{\text{n}}} \leq 7.39,\\
        1.18& \leq \frac{n_{\text{o}}}{n_{\text{n}}} \leq 1.34.
    \end{align}
    If at least two of the counts are non-zero and the corresponding above conditions of these non-zero counts are fulfilled, a molecule is scored as active, yielding a reward of 1. Otherwise, it is scored as inactive, giving a reward of 0. The noise model flips the score with probability $0.01$. 
\paragraph{Analyze}
    In each analyze step, the QSAR model of the scoring function is retrained using the previous and new test scores. Also, the new test scores are used to update the total reward and the number of plays of each ball in the Zooming algorithm, as described in Section \ref{sec:zooming}.

\subsection{Selection}
\label{sec:select_strategies}
In each cycle, $K=100$ molecules are selected to be made. We restrict us to the realistic scenario where previously selected molecules, of both the current and previous cycles, can not be selected again. This gives the extreme case of volatile base arms where no arm can be selected twice.
 
 In Section \ref{sec:zooming} we introduced our Zooming method with multiple plays and volatile arms, including both the weighted and unweighted versions. We use the doubling trick with no restart to incrementally increase the time horizon $T$. We use geometric doubling where the time horizon is defined by $T_i = 2^i$ for phases $i=0, 1,2,3,\dots$, such that each phase $i_{\mathrm{ph}}$ consists of $2^{i_{\mathrm{ph}}}-2^{i_{\mathrm{ph}}-1}$ rounds (except the first phase which is played for one round). For the sake of brevity, this is not included in Algorithm \ref{alg:experimental_procedure}. As noted by \textcite{besson2018doubling}, using the doubling trick with no restart is just a heuristic and it is difficult to state any theoretical results on this heuristic, but it can enjoy better empirical performance. In our problem, we have a limited number of total rounds, due to time and budget constraints, and we may not know the total number of rounds beforehand. By using the doubling trick with no restart, we do not discard any of the expensive information that has been acquired, while allowing the time horizon to be unknown before the start of the algorithm.

Below, we briefly describe the other strategies that we consider in the experiments: greedy selection and decaying-epsilon-greedy selection. In our experiments, we also investigate random sampling, which randomly (without replacement) selects base arms in $\mathcal{M}^t$.

\paragraph{Greedy selection}
In each cycle $t$, there are scores $\mathcal{F}^t$ from the scoring function. Greedy selection selects the $K$ distinct molecules with the highest scores. Ties are broken arbitrarily.

\paragraph{Decaying-epsilon-greedy selection}
In each round $t$, decaying-epsilon-greedy ($\epsilon_t$-greedy in short) selects a random compound with probability 
\begin{equation}
    \epsilon_t = \epsilon_{\text{min}} + \left(\epsilon_{\text{max}}-\epsilon_{\text{min}}\right)e^{c_{d}(t-1)},
\end{equation}
or the (non-selected) highest scoring molecule, according to $\mathcal{F}^t$, with probability $1-\epsilon_t$. This is done until $K$ base arms are selected. We use $c_{d} = 0.015$ and $\epsilon_{\text{min}} = 0.0$, and have investigated both $\epsilon_{\text{max}} = 0.6$ and $\epsilon_{\text{max}} = 0.4$.

\subsection{Comparison of Selection Strategies} 
The normalized cumulative rewards averaged over 10 runs are shown in \figurename~\ref{fig:cum_rew} for various selection strategies. 
Weighted Zooming, random sampling and greedy selection show similar (normalized) cumulative reward after 200 cycles. In the last cycles, they show significantly higher cumulative rewards compared to the other strategies. It is reasonable that greedy selection is able to select active molecules in the last cycles since then the scoring function has learned sufficient information about the ground truth to guide REINVENT towards generating active molecules.
The same is true with random sampling since it reflects the overall activity (reward) of the generated molecules. On the other hand, unweighted Zooming shows significantly lower cumulative rewards compared to the other strategies.  An explanation is that more exploration of the chemical space is needed to learn to identify the most rewarding areas since unweighted Zooming is not inclined toward selecting only actives, as is the case for weighted Zooming. 

\figurename~\ref{fig:novelty} shows the novelty of the selected actives averaged over 10 runs for each selection strategy.
Note that no actives are selected in some early cycles of certain runs and, therefore, these runs are excluded when computing the averages and confidence intervals. The novelty at cycle (round) $t$ is defined as the average dissimilarity to active molecules that have been selected in previous cycles 
\begin{equation}
    \texttt{novelty}_t = \sum_{m \in \cup_{i=1}^{t-1} \mathcal{A}^i} \frac{1}{|\cup_{i=1}^{t-1} \mathcal{A}_i|} \sum_{m' \in \mathcal{A}^t} \frac{\mathcal{D}\left(x_{m}, x_{m'}\right)}{|\mathcal{A}^t|} ,
\end{equation}
where $\mathcal{D}(\cdot,\cdot)$ is the Jaccard distance and $\mathcal{A}^t$ is the set of true actives selected in cycle $t$. That is, a novelty of 1 corresponds to selecting actives that are entirely dissimilar to previously selected actives, i.e., a completely novel selection.
In the first cycles, all strategies display a high novelty in the selection of actives. Overall, the novelties decrease as more cycles are performed, until after around 100 cycles when the average novelties converge to different values. After this point, unweighted Zooming shows significantly higher novelty compared to the other selection strategies. This observation suggests that unweighted Zooming is able to explore different areas of the chemical space more effectively. Weighted Zooming yields the second-best average novelty, while random sampling leads to the lowest average novelty. The fact that random sampling displays the lowest novelty indicates that the overall diversity of the generated molecules is low by this method, and possibly in general.

The mean of normalized cumulative reward and novelty $\left(\texttt{normalized\_cumulative\_reward}_t + \texttt{novelty}_t\right)/2$ for each cycle $t$, averaged over 10 runs, is shown in \figurename~\ref{fig:mean_rew_nov} for each strategy (for illustrative purposes, we discard the confidence intervals).
This indicates how the strategies handles the trade-off between maximizing the cumulative reward and selecting novel actives. For all strategies, the mean decreases in the beginning but then starts to increase after around 25 to 75 cycles, depending on the strategy. $\epsilon_t$-greedy with $\epsilon_{\text{max}} = 0.6$ and unweighted Zooming yield the largest mean for the first 50 cycles, while weighted Zooming and greedy selection show the largest means for at least the last 100 cycles. Hence, unweighted Zooming is a good choice in the early phase of the autonomous drug design, while weighted Zooming is better in the late phase. On the other hand, weighted Zooming still performs better than greedy selection in the early cycles.
This means that a strategy utilizing the benefits of both unweighted and weighted Zooming could possibly provide a promising way to handle the trade-off in all cycles.

\section{Conclusions}\label{sec:conclusion}
We have formulated the problem of determining what molecule to make next, in an autonomous drug design process, as a stochastic multi-armed bandit problem. To solve this problem, we adapted the contextual Zooming algorithm \cite{slivkins2011contextual} to a setting with multiple plays and volatile arms leading to two variants called weighted Zooming and unweighted Zooming.  
We compared our methods with random sampling, greedy selection and decaying-epsilon-greedy selection. After 200 cycles of a simulated DMTA cycle, we find out that Zooming performs overall best and has the potential to be used in autonomous drug design to handle the trade-off between selecting active (high rewarding) molecules and selecting structurally different but still active molecules.

\section*{Acknowledgements}
This work was partially supported by the Wallenberg Artificial Intelligence, Autonomous Systems, and Software Program (WASP), funded by the Knut and Alice Wallenberg Foundation, Sweden.

\printbibliography

\end{document}